\documentclass{article}

\usepackage{microtype}
\usepackage{graphicx}
\usepackage{subfigure}
\usepackage{booktabs} % for professional tables

\usepackage{hyperref}

\usepackage[accepted]{icml2023}

\usepackage{amsmath}
\usepackage{amssymb}
\usepackage{mathtools}
\usepackage{amsthm}

\usepackage[capitalize,noabbrev]{cleveref}

\theoremstyle{plain}
\newtheorem{theorem}{Theorem}[section]

\theoremstyle{definition}

\theoremstyle{remark}

\usepackage[textsize=tiny]{todonotes}

\usepackage{hyperref}
\usepackage{url}
\usepackage{microtype}
\usepackage{graphicx}
\usepackage{amsmath}
\usepackage{amsthm}
\usepackage{booktabs}
\usepackage{algorithm}
\usepackage{algorithmic}
\usepackage{wrapfig}
\usepackage{color}
\usepackage{bm}
\usepackage{xr}
\usepackage{multirow}
\usepackage{enumitem}

\usepackage{footnote}

\usepackage{amsmath}
\usepackage{amssymb}

\icmltitlerunning{A Study on ReLU and Softmax in Transformer}

\begin{document}

\twocolumn[
\icmltitle{A Study on ReLU and Softmax in Transformer}



\icmlsetsymbol{equal}{*}

\begin{icmlauthorlist}
\icmlauthor{Kai Shen}{equal,zju}
\icmlauthor{Junliang Guo}{equal,msra}
\icmlauthor{Xu Tan}{msra}
\icmlauthor{Siliang Tang}{zju}
\icmlauthor{Rui Wang}{msra}
\icmlauthor{Jiang Bian}{msra}
\end{icmlauthorlist}

\icmlaffiliation{zju}{Department of Computer and Science, Zhejiang University, \{shenkai, siliang\}@zju.edu.cn}
\icmlaffiliation{msra}{Microsoft Research Asia, \{junliangguo, xuta, ruiwa, jiabia\}@microsoft.com}

\icmlcorrespondingauthor{Xu Tan}{xuta@microsoft.com}

\icmlkeywords{Machine Learning, ICML}

\vskip 0.3in
]


\printAffiliationsAndNotice{\icmlEqualContribution}

\begin{abstract}
The Transformer architecture consists of self-attention and feed-forward networks (FFNs) which can be viewed as key-value memories according to previous works.
However, FFN and traditional memory utilize different activation functions (i.e., ReLU and Softmax respectively), which makes them not equivalent.
In this paper, we first rebuild the connections between FFN and key-value memory by conducting extensive studies on ReLU and Softmax, and find they are equivalent when adding an additional layer normalization module on Softmax.
In addition, ReLU outperforms Softmax on both FFN and key-value memory when the number of value slots is large.
We analyze the reasons and then explore this good property of ReLU on the self-attention network where the original Softmax activation performs poorly on long input sequences. 
We then propose a full ReLU architecture named ReLUFormer which performs better than the baseline Transformer on long sequence tasks such as document translation.
This paper sheds light on the following points: 
1) Softmax and ReLU use different normalization methods over elements which lead to different variances of results, and ReLU is good at dealing with a large number of key-value slots; 2) FFN and key-value memory are equivalent, and thus the Transformer can be viewed as a memory network where FFNs and self-attention networks are both key-value memories.

\end{abstract}

\section{Introduction}
\label{sec:intro}

Transformer~\cite{vaswani2017attention} models have achieved great success in various natural language processing tasks including large-scale language model pretraining~\cite{devlin2018bert,brown2020language} and machine translation~\cite{vaswani2017attention,hassan2018achieving}. A lot of works have been conducted to explore, analyze and explain the architecture. A series of works~\cite{dai2021knowledge,sukhbaatar2019augmenting,yao2022kformer} investigate how Transformers understand and store a huge amount of knowledge from data, mainly by building the connections between the network modules (i.e., self-attention network (SAN) and feed-forward network (FFN)) and key-value memory according to the similarity in their formats. For example, they take the two layers in FFN as key and value parameters that store different patterns of knowledge~\citep{geva2020transformer}.

However, previous works have not considered the difference in the activation function, which plays an important role in neural networks. Specifically, traditional key-value memory~(and also SAN) usually utilizes Softmax to normalize the output of query and key in order to highlight important values, whereas FFN utilizes ReLU as the activation function which does not introduce normalization.
In this paper, we conduct extensive studies and revisit the connections between SAN, FFN, and key-value memory while taking activation functions into account.

Specifically, for FFN, we first replace ReLU with Softmax to keep consistent with the format of memory, but observe performance degradation because the activation results become too small to carry enough information after being normalized by Softmax.
We introduce an additional layer normalization module on Softmax results to adjust the variance and show the identity of FFN and key-value memory.
In addition, we explore their scalability as the hidden dimension~(i.e., the dimension of the inner hidden vector between two layers) of FFN is usually very large on large-scale models~\citep{shazeer2017outrageously}. By changing the total number of key-value slots, we find that ReLU performs better than Softmax when the number of slots is larger. 
We explore the reason by calculating the ratio of top scores among all activations and find that the activation weights are highly centralized in a small number of slots, thus insufficient to utilize the context information of other slots, while ReLU is able to alleviate this problem.

Given the superior performance of ReLU when scaling to a large number of value slots, we then explore how ReLU performs on SAN where Softmax may have a trouble modeling long-sequences~\citep{sun2022length}.
Unfortunately, directly alternating Softmax to ReLU does not converge. With theoretical and experimental analysis, we find that the variance of SAN results with ReLU activation grows with the length of the input sequence, and the dynamic variance will lead to an unstable training process.
Therefore, a variance reduction factor and regularization loss functions are introduced to solve this problem.
As a result, we make it possible to utilize ReLU on self-attention, which performs better than Softmax when dealing with long input sequences.

In summary, this paper provides insights into the difference and relationship between ReLU and Softmax activation functions in the Transformer architecture.
\begin{itemize}[leftmargin=*]
    \item Softmax provides exponential normalization over all value slots and therefore highlights a small number of them while neglecting others, which may cause performance degradation when the number of slots is large, e.g., FFN with large hidden dimensions or SAN with long input lengths. ReLU bypasses this problem but faces variance exploding which varies over different training samples.
    \item We revisit the relations between the FFN and key-value memory and find that they are equivalent when additional layer normalization is introduced.
    \item With the ReLU activation function in both SAN and FFN components, we propose a fully-ReLU Transformer architecture (ReLUFormer). We claim that our ReLUFormer can be viewed as an integration of key-value memory, with FFNs as global key-value memories and SANs as local memories. We evaluate the model on the task of long-document translation and verify the superiority of the full ReLU architecture over the Transformer baseline.
\end{itemize}

The rest of the paper is organized as follows. We introduce the backgrounds of FFN, SAN, and key-value memory in Section~\ref{sec:back}. And then we revisit the connections between FFN and key-value memory in Section~\ref{sec2:reveal_connection_ffn_neuralmemory}. 
We then explore the connections between the ReLU and Softmax in SAN, and compare the ReLU and Softmax in the task of long document translation with the proposed fully-ReLU architecture in Section~\ref{sec:relu-softmax-selfattn}. We summarize and provide insights on our findings in Section~\ref{sec:summ}.

\section{Background}
\label{sec:back}

\subsection{Feed-Forward Network and Key-Value Memory} 
\label{sec:back_ffn}
Transformer~\citep{vaswani2017attention} has achieved great success on natural language processing tasks, such as machine translation and language modeling. 
Recently, some works have been proposed to analyze the architecture and investigate the secrete of the success of Transformers.
Some works have revealed the relation between feed-forward network (FFN) and key-value memory~\citep{geva2020transformer,sukhbaatar2019augmenting}. They intuitively regard the FFN as key-value memory by unifying them in the formulation.

\paragraph{Feed-Forward Network} 
Formally, given an input sequence representation $X \in \mathbb{R}^{n \times d}$ where $n$ is the length and $d$ is the dimension, an FFN consists of two linear projections with a non-linear activation function shown as follows:
\begin{equation}
    H = \textrm{ReLU}(X \cdot W_1^T + b_1) \cdot W_2 + b_2,
\label{equ:ffn-relu}
\end{equation}
where $H$ is the output hidden representation, $W_1, W_2 \in \mathbb{R}^{d_h \times d}$ are learnable parameters, and $b_1, b_2$ indicate bias terms which are omitted in the following\footnote{We omit the bias terms since it contains few parameters and has little influence on the results~\cite{geva2020transformer}.}, $d_h$ is the hidden dimension of FFN.

\paragraph{Key-Value Memory} 
The key-value memory networks~\cite{SainbayarSukhbaatar2015EndToEndMN,geva2020transformer} consist of learnable key-value parameters, which are designed to store the knowledge in the training set.
Given an input query $X \in \mathbb{R}^{n \times d}$, the output is computed by aggregating the values $V \in \mathbb{R}^{d_h \times d}$ w.r.t the distribution computed by keys $K \in \mathbb{R}^{d_h \times d}$ as follows:
\begin{equation}
    H = \textrm{Softmax}(X \cdot K^T) \cdot V,
    \label{equ:kvm}
\end{equation}
where $d_h$ is the number of memory slots.

\paragraph{Relations} 
Previous works~\citep{sukhbaatar2019augmenting,geva2020transformer} reveal that the FFN and key-value memory are similar in the formulation, i.e., by regarding $W_1$ as keys and $W_2$ as values the FFN can be viewed as a kind of key-value memory.
In addition, \citet{dai2021knowledge} conducts experiments on how knowledge is stored in a pre-trained model and finds that there are some knowledge neurons in the FFN layer related to the expression of factual knowledge. 
\citet{lample2019large} introduces a large-scale external memory based on product keys and successfully integrates it into transformer architecture by replacing the FFN layer. However, there still exists a difference in the choice of activation functions, where the FFN usually adopts ReLU and the key-value memory uses Softmax, which may lead to different model performance. In this paper, we will explore the connections between FFN and key-value memory by studying the ReLU and Softmax.

\subsection{Self-Attention Network and Key-Value Memory}
\label{sec:back_self}
As for the self-attention network (SAN), which is initially proposed in a key-value computation format~\citep{vaswani2017attention}
previous works have also explored its relation with key-value memory~\citep{sukhbaatar2019augmenting}.
Formally, given the input sequence $X \in \mathbb{R}^{n \times d}$, the self-attention is calculated as follows:
\begin{equation}
        H = \textrm{Softmax}(\frac{ (X W_Q) \cdot (X W_K)^T}{\sqrt{d}} ) \cdot X W_V, 
\end{equation}
where $W_Q, W_K, W_V \in \mathbb{R}^{d \times d}$ are learnable parameters. By denoting $\hat{X} \coloneqq X W_Q / \sqrt{d}$, $\hat{K} \coloneqq X W_K$ and $\hat{V} \coloneqq X W_V$, the SAN is identical to key-value memory in Equation~(\ref{equ:kvm}) as well.
\citet{wu2022memorizing} introduces an external memory and integrates it with SAN by a gating mechanism. \citet{dai2019transformer} reuses the states of previous segments and augments them in SAN to capture long-term dependencies.
These works conceptually regard the SAN as a local memory, in which the query is the current token and the keys and values are other context tokens.

In conclusion, although FFN, SAN, and key-value memory are similar in formulation, previous works have not carefully discussed the differences in activation functions. In practice, it is a convention to use ReLU in FFN and Softmax in SAN and key-value memory. In this paper, we provide in-depth analyses of ReLU and Softmax as well as their performance on FFN and SAN. We start by revisiting the connections between FFN and key-value memory.

\section{Connections Between FFN and Key-Value Memory}
\label{sec2:reveal_connection_ffn_neuralmemory}

\subsection{ReLU and Softmax are Different}
We first investigate whether the difference in ReLU and Softmax activation functions will influence the performance of FFN and key-value memory.
We conduct a preliminary experiment that simply replaces FFN layers with key-value memories~(i.e., replacing Equation~(\ref{equ:ffn-relu}) with Equation~(\ref{equ:kvm})) in a vanilla Transformer and keeps other components the same. We test on the machine translation task and the IWSLT14 De-En benchmark dataset~(refer to Appendix~\ref{append:dataset_nmt} for more details). We report the BLEU score~\cite{papineni2002bleu} as the evaluation metric.

\begin{table}[tb]
\caption{The results on IWSLT14 De-En translation task with different activation functions on FFN. The variance ratio indicates the ratio between the variance of the FFN output and residual. Layer Norm denotes the layer normalization layer applied on the result of FFN.}
\label{table:method_ffn_competitionandscale}
\vskip 0.15in
\begin{center}
\begin{tabular}{llcc}
\toprule
Activation & Layer Norm & BLEU & Variance Ratio \\ 
\midrule
ReLU         & No       & $34.22$  &  $0.27$  \\ 
\midrule
Softmax       & No            & $33.08$ & $0.01$\\
Softmax       & Yes           & $34.21$  & $0.34$ \\
\bottomrule
\end{tabular}
\end{center}
\end{table}

As shown in Table~\ref{table:method_ffn_competitionandscale}, we find the model performance drops from $34.22$ to $33.08$ in the BLEU score.
According to the discussion in the previous section, the only difference lies in the activation function, we can conclude that the drop comes from changing ReLU to Softmax, which has a significant influence and can not be ignored.

\subsection{Bridge the Gap between FFN and Key-value Memory}
Then, we analyze the reason for the performance drop. 
When computing results, Softmax normalizes the score of all slots while ReLU does not. Intuitively, the results of Softmax will have a much smaller variance than the results of ReLU.
In consequence, the residual from previous layers will dominate the output of the current FFN layer with the Softmax activation function, resulting in inefficient utilization of parameters and degeneration of model capacity.
To demonstrate this phenomenon, we compute the ratio between the variance of FFN output and residual for both ReLU and Softmax. The results in Table~\ref{table:method_ffn_competitionandscale} verify our claim, where the ratio of output with Softmax is much smaller than that of ReLU. 

To alleviate this problem, we add a layer normalization~\cite{ba2016layer} module after the FFN layer to learn and adjust the variance ratio of the output, i.e., Equation~(\ref{equ:kvm}) becomes to $H = \textrm{LN}(\textrm{Softmax}(X \cdot K^T) \cdot V)$.
The experimental results are shown in Table \ref{table:method_ffn_competitionandscale}. With layer normalization, both the variance ratio and BLEU scores of Softmax are promoted, and the FFN with Softmax performs similarly to ReLU. Therefore, we amend the previous findings~\citep{geva2020transformer} that the FFN and key-value memory are equivalent when an additional layer normalization module is introduced.

\begin{figure}[tb]
  \vskip 0.2in
  \begin{center}
  \centerline{\includegraphics[width=1.0\linewidth]{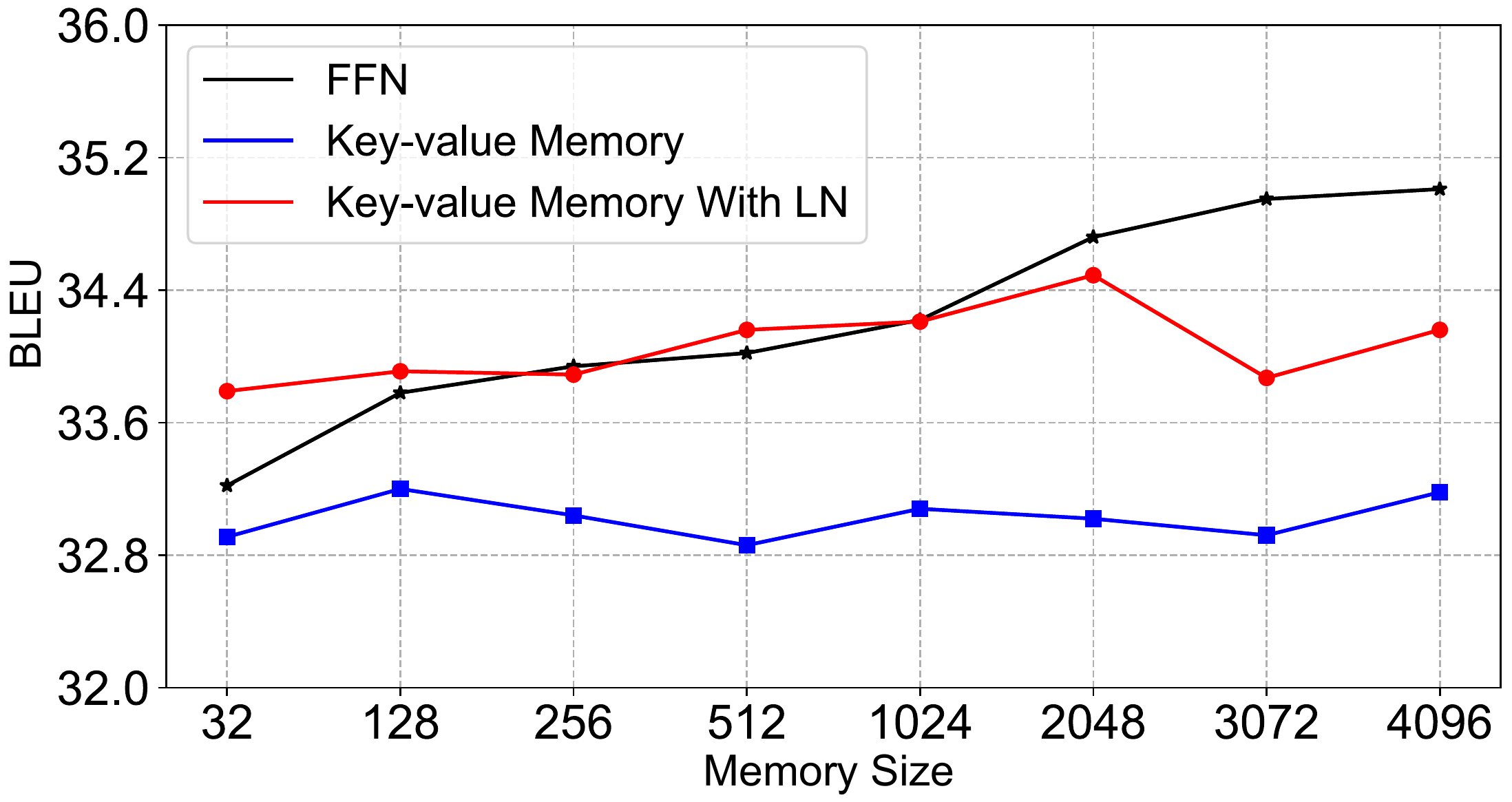}}
  \caption{The BLEU scores of FFN, key-value memory, and key-value memory with layer normalization (key-value memory with LN) on different memory sizes.}
  \label{fig:method_ffn_memory_longseq}
  \end{center}
  \vskip -0.2in
\end{figure}

\begin{figure*}[tb]
  \vskip 0.2in
  \begin{center}
  \centerline{\includegraphics[width=1.0\linewidth]{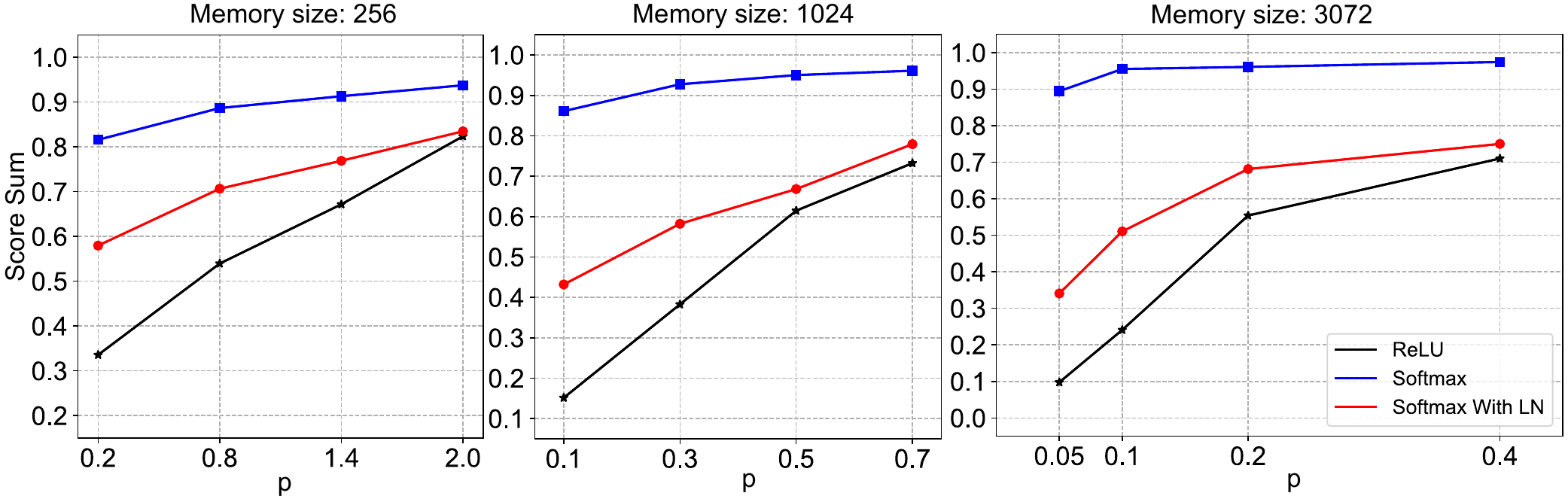}}
  \caption{The visualization of top-$p\%$ score sum of ReLU, Softmax, and Softmax with layer normalization (Softmax with LN).}
  \label{fig:ffn_toppvisualize}
  \end{center}
  \vskip -0.2in
\end{figure*}

\subsection{Scaling to Large Number of Values}
\label{sec:performance_on_large_number_of_values}
With the progress of large-scale Transformer networks, the hidden dimension (or the number of slots from the view of memories) $d_h$ of FFN layers becomes larger and brings better results as shown in previous works~\citep{shazeer2017outrageously,fedus2021switch}. To verify the performance and show the generalization ability of different activation functions when the number of values is large,
we vary $d_h$ from $32$ to $4096$ and train three models including FFN with ReLU~(ReLU), FFN with Softmax~(Softmax), and FFN with Softmax and layer normalization~(Softmax with LN).
Results are shown in Figure~\ref{fig:method_ffn_memory_longseq}. 
The observations are twofold. 1) The ReLU activation function is superior to Softmax consistently on all sizes. 2) When equipped with LN, Softmax performs comparably to ReLU. However, when the memory size is large (i.e., $3072$, $4096$), the ReLU performs better than Softmax with LN.
In conclusion, ReLU shows stronger capacity when dealing with a large number of values than Softmax.

\subsection{Quantitative Analysis between ReLU and Softmax}
\label{sec:ffn_quant_ana}

We conjecture the reason is the exponential normalization in Softmax. Concretely, since Softmax provides the exponential normalization on the elements while ReLU does not, 
Softmax provides over-centralized distribution over elements, which means only a few elements are highlighted while occupying most weights.
Then when the memory size is large, Softmax will overlook most value slots and only utilize a few of them, which does not benefit from the large size of memory. In contrast, there is no competition among elements in ReLU, which is able to aggregate more knowledge. A straightforward method to alleviate this problem is to increase the temperature in Softmax to flatten the output distribution. However, we empirically find it has little effect in experiments.

To verify the conjecture,
we qualitatively analyze the competition by visualizing the score distribution, i.e., $\textrm{Softmax}(X\cdot K^T)$ in Equation~(\ref{equ:kvm}). For each distribution, we sort scores in descending order and calculate the summation of the top-$p\%$ elements. We normalize ReLU scores to make sure the summation of all scores is $1$ as $\text{ReLU}(x_i k_j^T) / \sum_{p=1}^{n}\text{ReLU}(x_i k_p^T) \label{eq:relu_norm}$, where $x_i, k_j$ is the $i$-th, $j$-th elements of query $X$, key $K$ mentioned in Section~\ref{sec:back_ffn}.
Therefore, a higher top-$p\%$ score sum indicates a more centralized distribution, i.e., all scores are concentrated on a small number of elements. From Figure~\ref{fig:ffn_toppvisualize}, we can find that Softmax provides a highly centralized distribution as top $0.2\%$ elements occupy more than $85\%$ scores, and becomes more severe when the memory size grows. 
In contrast, ReLU can alleviate this problem and therefore utilize the information of more memory slots.

\begin{figure}[tb]
  \vskip 0.2in
  \begin{center}
  \centerline{\includegraphics[width=1.0\linewidth]{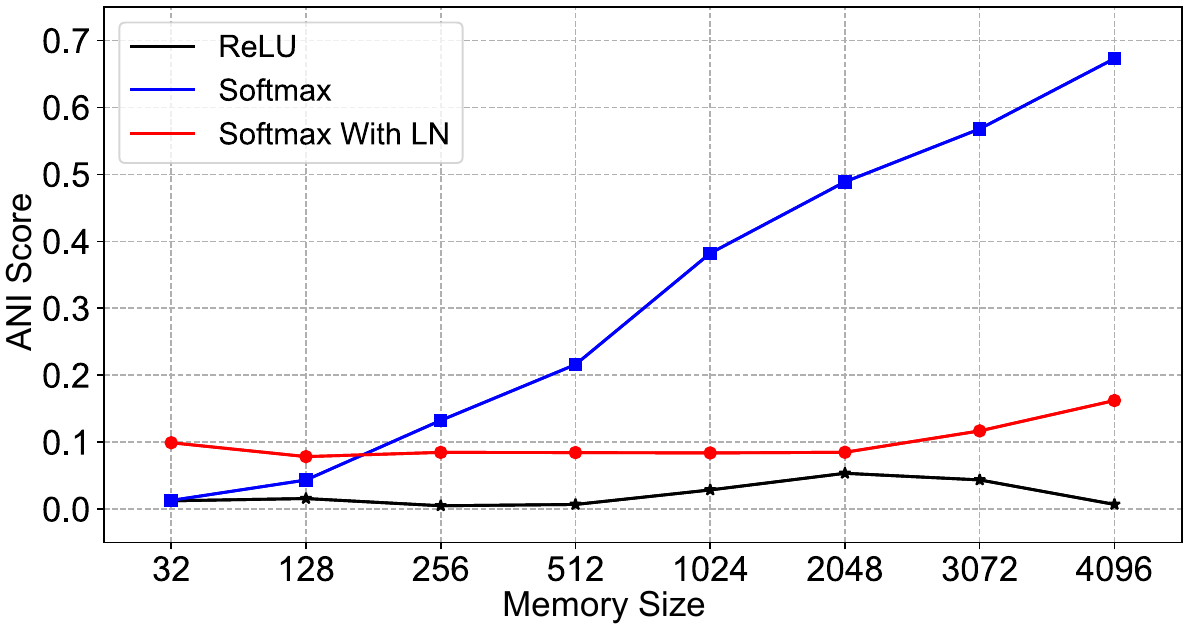}}
  \caption{The anisotropy (ANI) score of ReLU, Softmax, and Softmax with layer normalization (Softmax with LN).}
  \label{fig:ffn_diversity}
  \end{center}
  \vskip -0.2in
\end{figure}

As a result of the over-centralized distribution, the optimization of memory slots with Softmax will be sub-optimal as most of them cannot receive enough gradient due to small scores. We then quantitatively evaluate the quality of the learned values by measuring their anisotropy~\citep{ethayarajh2019contextual}, 
which is defined as the average of pair-wise similarities and therefore the lower the better, i.e., value slots are different from each other and able to contain discriminative and diverse information. The anisotropy score (ANI) is defined as follows:
\begin{equation}
    \textrm{ANI} = \frac{1}{d_h \cdot (d_h -1)}{\sum_{i=1}^{d_h} \sum_{j=1, j \neq i}^{d_h} \frac{V_i^T V_j}{\| V_i \| \cdot \| V_j \|} },
\end{equation}
where $V_i$ indicates the $i$-th value slot. 
Visualization results are illustrated in Figure~\ref{fig:ffn_diversity}, from which we can find that the values of Softmax collapse and fail to store diverse knowledge, and adding LN alleviates this problem while utilizing ReLU achieves the best performance.

\subsection{Summary}
With these explorations, we have the following insights. 
\begin{itemize}[leftmargin=*]
    \item The results of Softmax and ReLU have different properties, including variance and normalization.
    For variance, the ReLU has a larger variance compared with Softmax, which is more expressive. The hidden output layer with a small variance may be dominated by the residual during end-to-end training, which can lead to the waste of the parameters and sub-optimal results.
    For normalization, since Softmax provides exponential normalization on the elements while ReLU does not, the distribution of Softmax is more centralized.
    When the memory size is large, Softmax will overlook most of the elements, thus resulting in a less diverse and discriminative memory value space.
    \item When the Softmax is equipped with layer normalization in key-value memory, the FFN and key-value memory can be equivalent. The layer normalization can largely alleviate the small variance and over-centralized distribution brought by Softmax in the key-value memory, thus the Softmax with layer normalization can achieve comparable performance with FFN.
    \item We find that the ReLU is more capable to deal with a large number of memory slots. When the number of memory slots is larger, the distribution of Softmax is more centralized, which results in the inefficient utilization of the memory slots. 
\end{itemize}
The last observation also inspires us that when handling long sequences in self-attention, it is beneficial to pay attention to more elements instead of centralizing on a small portal. However, since self-attention is similar to key-value memory and it is conventional to use Softmax as the activation function, will ReLU perform better than Softmax when handling long sequences? We will explore the differences of ReLU and Softmax in self-attention in the next section.

\section{ReLU vs Softmax in Self-Attention}
\label{sec:relu-softmax-selfattn}

Given the findings that ReLU outperforms Softmax in FFN, when dealing with a large number of value slots, a natural question is how will ReLU perform on the self-attention network (SAN). As discussed in Section~\ref{sec:back_self}, SAN can be straightforwardly formatted as key-value memory, with queries, keys, and values as different representations of the input. Then the memory size is denoted by the length $n$ of the input instead of the hidden dimension $d_h$ in FFN. Therefore, we expect to observe the superiority of ReLU over Softmax when dealing with long sequences, following the conclusions of the previous section. 

Similarly, we conduct preliminary experiments by directly replacing Softmax with ReLU, but we find the model fails to converge. Specifically, we find the variance of SAN results exploding. To solve the variance exploding problem, we add a layer normalization layer succeeding to the SAN results to adjust the variance of SAN. Unfortunately, there still occurs performance degradation.
Such phenomenon is also reported in previous studies~\cite{zhang2021sparse}. 
Therefore, in this section, we first analyze the reasons that ReLU fails and then propose our solutions. 
We then compare the performance of ReLU and Softmax on different sequence lengths.

\subsection{Solving Variance Exploding Caused by ReLU}
Recall the formulation of SAN with ReLU activation:
\begin{equation}
    h_i = \sum_{j=1}^{n}{\text{ReLU}(q_i^Tk_j)v_j}, 
\label{equ:sec4-relu}
\end{equation}
where $q_i, k_j, v_j$ is the $i,j,j$-th element of query $\hat{X}$, key $\hat{K}$, and value $\hat{V}$ mentioned in Section~\ref{sec:back_self} respectively, $h_i$ is the output representation of the $i$-th token, and $n$ is the sequence length. We find that the variance of $h_i$ is dependent on the sequence length $n$ by the following theory introduced by \citealp{he2015delving}:

\begin{theorem} 
\label{theorem:section4}
Given $n$ random variables $x_i \sim \mathcal{N}(0, 1), i\in [1, n]$ and $v_j \sim \mathcal{N}(0, 1), j \in [1, n]$, $y_i$ defined as:
\begin{equation}
    y_i = \sum_{i=1}^{n} {\textrm{ReLU}(x_i)v_i}. \nonumber
\end{equation}
Then $y_i$ follows Gaussian distribution $\mathcal{N}(0, \frac{n}{2})$.
\end{theorem}

Based on Theorem~\ref{theorem:section4}, in Equation~(\ref{equ:sec4-relu}), the output $h_i$ will follow the distribution $N(0, \frac{n}{2})$. Therefore, the variance of results grows with the sequence length, and directly replacing Softmax with ReLU will lead to instability of the training process.
Empirically, although adding layer normalization is supposed to learn and re-scale the variance, we find it still leads to
sub-optimal BLEU results. We conjecture the different performance of LN on FFN and SAN is due to the dynamic memory size in SAN (i.e., the sequence length $n$) which is static in FFN (i.e., the hidden dimension $d_h$), and thus the LN module is not able to learn appropriate variance for sentences with different lengths.

\begin{table}[tb]
\caption{The BLEU scores on IWSLT14 De-En task with different activation functions on self-attention. w/ Scale Factor indicates whether the self-attention network has the scale factor. '-' indicates the setting does not converge. }
\vskip 0.15in
\begin{center}
\begin{tabular}{llc}
\toprule
Activation & w/ Scale Factor & BLEU  \\ 
\midrule
Softmax      & No   & $34.22$ \\ 
ReLU     & No   & -     \\ 
ReLU   & Yes    & $33.19$   \\
\bottomrule
\end{tabular}
\end{center}
\label{table:method_selfattn_relu}
\end{table}

Motivated by these analyses, 
to stabilize the variance of ReLU output, 
% to solve the scale problem, 
we propose the variance reduction factor defined as $\gamma \sqrt{n/2}$, where $\gamma$ is a hyper-parameter. This is similar to the implementation of Kaiming Normalization~\cite{he2015delving} but we adapt it during end-to-end training instead of initialization since the sequence length $n$ is dynamic.
Formally, the Equation~\ref{equ:sec4-relu} goes to:
\begin{equation}
    h_i = \sum_{j=1}^{n}{\frac{\text{ReLU}(q_i^Tk_j)}{\gamma \sqrt{n/2}} v_j}. 
\label{equ:sec4-relu-scale}
\end{equation}
Then we apply it to the SAN and obtain $33.19$ BLEU scores on IWSLT14 De-En machine translation task as shown in Table~\ref{table:method_selfattn_relu}. 
Although this is a big step towards a successful ReLU-based self-attention model, it still has a gap of $1.03$ BLEU scores to the vanilla Softmax-based self-attention.
We then go a step further and close the gap between ReLU and Softmax for SAN in the following section.

\subsection{Closing the Gap Between ReLU and Softmax}
\label{sec:slf-attn-quantitiveanalysis}

To better analyze the performance gap between ReLU and Softmax-based SAN, we denote the weight distribution over values as $s=(s_1, ..., s_n)$ where $s_j = \textrm{ReLU}(q_i^T k_j)/ \gamma \sqrt{n/2}$ for ReLU w/ variance reduction, and $s_j = \textrm{Softmax} (q_i^T k_j)$ for Softmax, and then compute the entropy of $s$, i.e., $H(s) = - \sum_{i=1}^{n}{s_i \log(s_i)}$. We normalize the output of ReLU to ensure the summation is $1$ with the same method used in Section~\ref{sec:ffn_quant_ana} as $\text{ReLU}(s_i) / \sum_{j=1}^{n}\text{ReLU}(s_j)$.
Theoretically, a larger entropy indicates the distribution is more uniform, while a smaller one indicates the distribution is more centralized.
And in our case, we want to find a balance where the entropy is neither too small nor too large, i.e., the weight distribution is not too uniform or over-centralized. In this way, the context information can be well utilized.

The computed entropy results are listed in Table~\ref{table:quantitive_slfattn_relu_and_softmaxscores}.
The distribution learned with ReLU has a very small entropy, and $94\%$ weights are all zeros. Therefore, the ReLU on self-attention leads to weight distributions that are too sparse to cover enough context.
To alleviate this problem, we propose a regularization loss including two parts.
Firstly, we introduce a normalization regularization to enlarge the weights. Secondly, we constrain the entropy of the learned distribution by an entropy-margin regularization to make it more informative. The loss functions are shown as follows:
\begin{equation}
\begin{split}
 \mathcal{L}_{\textrm{reg}} = \log(\sum_{i=1}^{n}{s_i}) + \max(H(s) - C, 0)
\end{split}
\label{equ:normalization_loss}
\end{equation}
where $|\cdot|$ indicates the absolute value, $C$ is a constant that represents the upper bound of $H(s)$. The normalization regularization encourages the summation of weights $\sum_{i=1}^{n}{s_i}$ to be $1$, and therefore results in more non-zero elements. Note that different from the normalization in Softmax, it is not a hard constraint and therefore we do not observe the over-centralized distribution as that in Softmax. In contrast, the distribution becomes flat as we find the entropy grows drastically after the loss function is added. The entropy-margin regularization will encourage the model to try to keep the entropy of weight distribution $s_i$ smaller than the upper bound $C$, where $C$ is a hyper-parameter which we describe in the Appendix \ref{append:dataset_nmt}.

\begin{table}[tb]
    \caption{The entropy of the weight distribution $s$ with Softmax and ReLU activation functions on the IWSLT14 De-En test set.}
    \vskip 0.15in
    \begin{center}
    \begin{tabular}{lc}
    \toprule
   Activation   & $H(s)$ \\ 
   \midrule
Softmax & 1.40 \\
ReLU & 3.45 \\
\bottomrule
\end{tabular} 
    \end{center}
\label{table:quantitive_slfattn_relu_and_softmaxscores}
    % \vspace{-15 pt}
\end{table}

It is worth noting that the Transformer contains casual self-attention and cross-attention in the decoder, which are slightly different from the self-attention we have discussed aforementioned. To generalize the ReLU-based SAN to the decoder, we propose two solutions. 1) For the causal self-attention, we assign different lengths to each token as tokens after the current one are masked off.
2) The cross-attention does not require additional adaptations thus we treat it as the self-attention in the encoder. By packing all the components, we propose a fully ReLU Transformer named ReLUFormer.

\begin{table*}[tb]
\caption{The experimental results of sentence-level translation} on IWSLT14 De-En and WMT14 En-De datasets. $\Delta$Speed: relative translation speed compared with vanilla Transformer on WMT14 test set. Higher speedup indicates better efficiency.
\vskip 0.15in
\begin{center}
\begin{tabular}{lccc}
\toprule
                    & IWSLT14 De-En              & WMT14 En-De & $\Delta$Speed         \\ \hline
Vanilla Transformer~\cite{vaswani2017attention} & $34.22$ & $27.21$  & $1.00 \times$ \\
Sparsemax~\cite{martins2016softmax}           &       $33.98$                     &    $27.32$    & $0.56 \times$                  \\
1.5Entmax~\cite{peters2019sparse} & $34.46$ & $27.11$ & $0.52 \times$  \\
ReLA~\cite{zhang2021sparse}         & $33.59$ & $26.31$ & $0.99 \times$  \\ \hline
ReLUFormer          & $34.56$ & $27.64$ & $1.01 \times$  \\ \bottomrule
\end{tabular}
\end{center}
\label{table:conventional_nmt_exps}
\end{table*}

\subsection{ReLUFormer and Its Performance on Translation}
In this section, we will first demonstrate the effectiveness of the proposed ReLUFormer on traditional sentence-level machine translation benchmarks. Then, we compare ReLU with Softmax when dealing with long sequences on document-level benchmarks.

\subsubsection{Experiments on Sentence-Level Translation}
We evaluate ReLUFormer on the sentence-level translation task, a seminal task in NLP. 
We consider two benchmark
machine translation datasets, i.e., IWSLT14 German-English and WMT14 English-German. Details of datasets can be found in Appendix \ref{append:dataset_nmt}.

\paragraph{Baselines}
In addition to the vanilla Transformer~\citep{vaswani2017attention}, we also consider other sparse activation baselines which alternate Softmax with other functions: 1) Sparsemax~\cite{martins2016softmax}, 2) 1.5Entmax~\cite{peters2019sparse}, and 3) Rectified Linear Attention (ReLA)~\cite{zhang2021sparse}. The Sparsemax and 1.5Entmax are specially designed sparse attention activation functions similar to ReLU. And ReLA is another baseline simply replacing Softmax with ReLU, in which they propose the RMS normalization mechanism to address the variance problem caused by ReLU. We leave the detailed introduction of baselines to Appendix \ref{appendix:baseline_details_nmt}.

\paragraph{Results}
The results of ReLUFormer and baselines are listed in Table~\ref{table:conventional_nmt_exps}, from which we have 
the following observations. 1) Our proposed ReLUFormer outperforms the vanilla Transformer baseline, showing the effectiveness of the proposed techniques. 
2) When comparing with sparse attention-based baselines, our model also achieves consistent improvements over the Sparsemax, 1.5Entmax, and ReLA baselines by a large margin.
3) By comparing the latency during inference, we first find that ReLUFormer is comparable with the vanilla Transformer, while slightly faster than the ReLA and at 1.7 times faster than the Sparsemax and 1.5Entmax methods. 

\begin{table}[tb]
\caption{The ablation study on IWSLT14 De-En task. The reg loss indicates the regularization loss. ``-" denotes the model can not converge.}
\vskip 0.15in
\begin{center}
\begin{tabular}{lcc}
\toprule
 & BLEU & Entropy \\
\midrule
ReLUFormer     & $34.56$      & $1.60$  \\
- scale factor   & -   & - \\
- reg loss & $33.19$ & $0.08$  \\
\bottomrule
\end{tabular}
\end{center}
\label{table:nmt_ablation}
\end{table}

\paragraph{Ablation Study}
\label{sec:abl}
In this section,  we conduct ablation studies on the proposed ReLUFormer to further discuss the effectiveness of the proposed three techniques: 1) attention scale factor (scale factor), and 2) regularization loss (reg loss). We demonstrate their effectiveness by removing each part individually. Besides the BLEU scores, we also use the entropy mentioned in Section~\ref{sec:slf-attn-quantitiveanalysis} to quantitatively analyze the quality of the weight distribution.

The results are shown in Table~\ref{table:nmt_ablation}, and the observations are as follows. 1) The model cannot converge by removing the scale factor because of the exploding of the variance, and adding the variance reduction factor stabilizes the training of the model. 2) By removing the normalization loss, the performance drops by 1.37 BLEU and the entropy decreases by 1.52.  It illustrates that the regularization loss can provide a more informative attention distribution.

\subsubsection{Experiments on Document-Level Translation}

In this section, to verify our findings in Section~\ref{sec2:reveal_connection_ffn_neuralmemory} that ReLU performs better than Softmax when the number of value slots is large, we conduct experiments on the document-level translation task with long input and output sequences, because the length of the sequence represents the number of local memory slots in self-attention networks. We construct the long documents from a widely used document translation dataset Europarl7 En-De~\citep{maruf2019selective,zheng2020towards}. To compare our method under different lengths of input sequences, we reconstruct the documents by concatenating sentences to different limitations of lengths. The details can be found in Appendix \ref{appendix:document_translation}. We conduct our experiment on 5 generated datasets with length limit $\{128, 256, 512, 1024, 2048\}$.

\paragraph{Results}
We compare the proposed ReLUFormer with the vanilla Transformer~\citep{vaswani2017attention} and the Sparsemax~\citep{martins2016softmax} baselines in document neural translation with different lengths. The results are shown in Table~\ref{table:document_nmt}. 

We have the following observations. 1) We find that when the sequence length is small (i.e., 128, 256), the ReLUFormer is comparable with the vanilla transformer and the Sparsemax baseline, showing that both methods are able to deal with relatively smaller lengths. 
2) When the sequence length is large (i.e., 512, 1024, 2048), we find that our ReLUFormer consistently outperforms the vanilla transformer and the Sparsemax baselines. 
For example, when the sequence length is 1024 in Europarl7, ReLUFormer achieves 1.15 BLEU gains in translation quality. And the Sparsemax fails to converge when facing long sequences. It confirms that when the sequence is long, the ReLU is more effective.

\begin{table}[tb]
\caption{The experimental results of document translation with different document lengths on the Europarl7 dataset.}
\vskip 0.15in
\begin{center}
\begin{tabular}{lccccc}  
\toprule
                    & \multicolumn{5}{c}{Sequence Length}   \\
                    & $128$ & $256$   & $512$   & $1024$  & $2048$  \\ \midrule
Transformer &     $31.19$      & $31.94$ & $31.81$ & $31.74$ & $31.89$  \\
Sparsemax & $31.04$ & $31.70$ & - & - & -  \\ \midrule
ReLUFormer         &  $31.22$  & $32.09$ & $32.69$ & $32.89$ & $32.58$   \\
\bottomrule
\end{tabular}
\end{center}
\label{table:document_nmt}
\end{table}

\begin{figure}[tb]
  \vskip 0.2in
  \begin{center}
  \centerline{\includegraphics[width=0.9\linewidth]{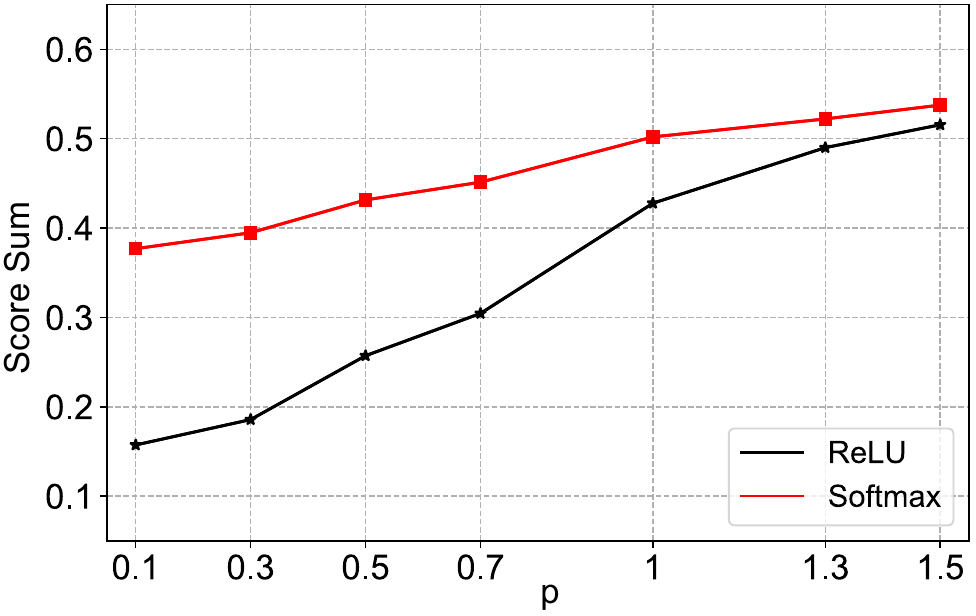}}
  \caption{The visualization of top-p\% score sum.}
  \label{fig:self-attn-topkscores}
  \end{center}
  \vskip -0.2in
\end{figure}

\begin{figure}[tb]
  \vskip 0.2in
  \begin{center}
  \centerline{\includegraphics[width=1.0\linewidth]{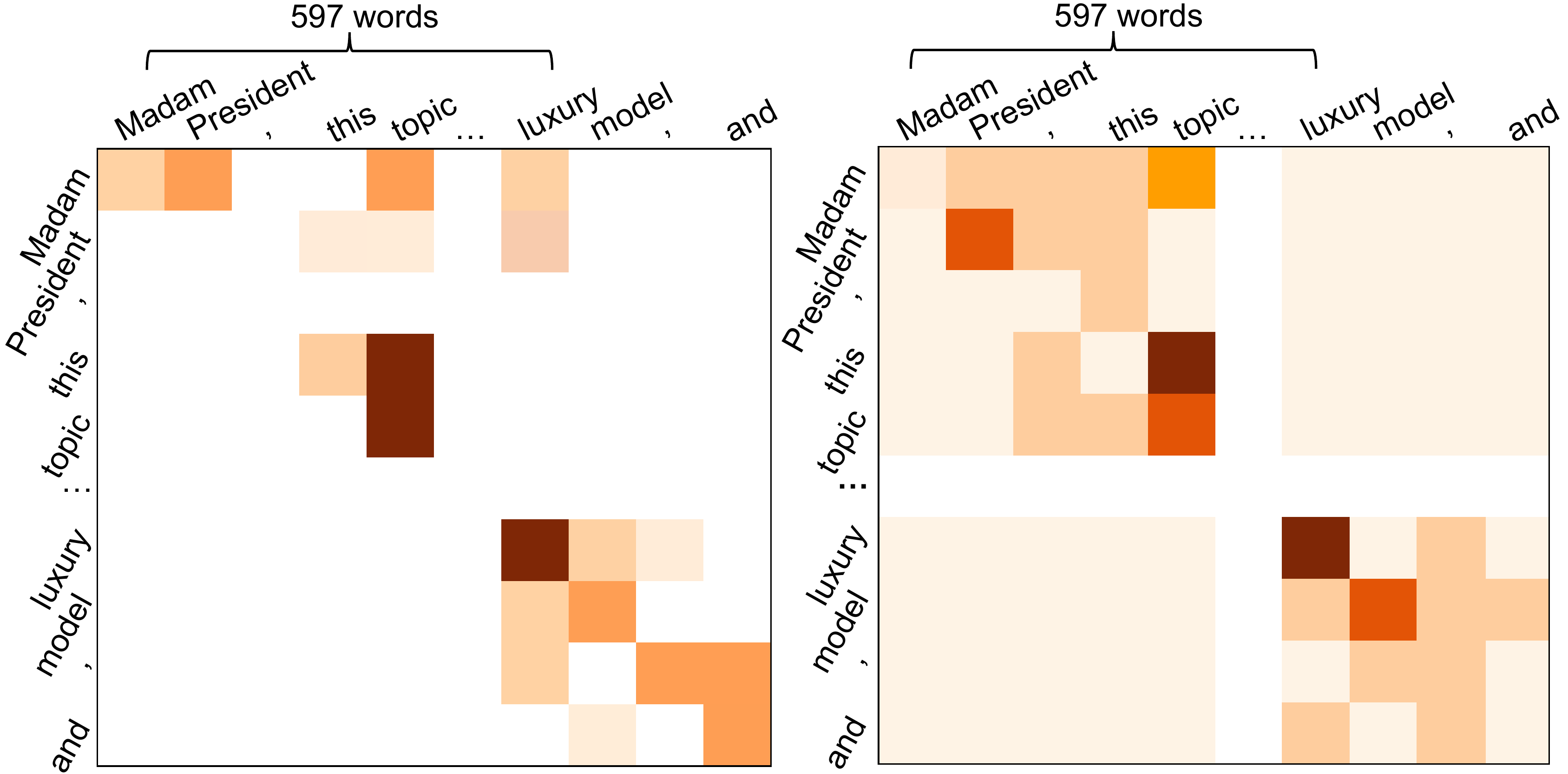}}
  \caption{The visualization of self-attention for ReLU (left) and Softmax (right). The case is from Europarl7 test set with $1024$ length limit.}
  \label{fig:attn_vis}
  \end{center}
  \vskip -0.2in
\end{figure}

\paragraph{Quantitative Analysis}
In this section, we intuitively explain why ReLU outperforms Softmax in long document translation. Similar to the study in FFN and key-value memory, we also visualize the top-$p$\% elements mentioned in Section~\ref{sec:ffn_quant_ana} of the activated scores for both Softmax and ReLU activation functions from the Europarl7 with 1024 length. A higher top-p\% score sum indicates a more centralized distribution. 
In practice, we observe that the specific token rating in top-1.5\% only occupies 2.1\% and 4.4\% scores for Softmax and ReLU, respectively. It means the tokens rating before top-1.5\% dominate the quality of the performance. Thus we only visualize the elements rating before top-1.5\%.
From Figure \ref{fig:self-attn-topkscores}, we can find that the Softmax provides a more centralized distribution compared with ReLU, which is similar to the observation in Section~\ref{sec:performance_on_large_number_of_values}. When modeling long sequence inputs in self-attention, the over-centralized distribution will pay attention to fewer contexts, which results in sub-optimal performance.

We also provide self-attention visualization to further demonstrate the superiority of ReLU. We randomly select a case in the Europarl7 test set with 1024 length and visualize the attention map in Figure \ref{fig:attn_vis} for both ReLU and Softmax activation functions.  
We have the following observations. 
1) We observe that the ReLU can capture more distant correlations. For example, the word ``Madam" and ``President" have relatively large attention weights in ReLU while small weights in Softmax. 
It shows the ReLU can capture more distant correlations compared to Softmax, which is beneficial to long sequence modeling.
2) We observe that the ReLU has less noise than Softmax. The ReLU assigns smaller attention values to some stop words such as ``," and ``this". Since the stop words contain little contextual information, it shows the ReLU has less noise.

\subsection{Summary}

In this section, we are motivated to explore whether ReLU outperforms Softmax in self-attention when handling long sequences. 
\begin{itemize}[leftmargin=*]
    \item We find that the variance of SAN results produced by ReLU is dependent on sequence length and therefore dynamic. Thus when ReLU is directly applied to replace Softmax in SAN, it can cause variance exploding and training instability. 
    \item With the extensive analysis of the performance degradation of ReLU, we propose solutions correspondingly to make the ReLU performs competitively to Softmax in SAN on the sentence-level translation task.
    \item Similar to the observation in Section~\ref{sec2:reveal_connection_ffn_neuralmemory}, we also find that the Softmax tends to generate a more centralized distribution which restricts the utilization of more context information especially when the sequence is long, while ReLU does not have the restriction.
\end{itemize}
By applying the ReLU-based Transformer to the long document translation task, we verify that the ReLU outperforms Softmax when the input sequence is long. 

\section{Insights and Findings}
\label{sec:summ}
We summarize the insights and findings of the paper in this section.
\begin{itemize}[leftmargin=*]
    \item Softmax and ReLU are different in Transformer from the perspective of variance and normalization. 
    1) Regarding the variance, the result of ReLU has a larger variance compared with that of Softmax. 
    The hidden representation with a small variance will be dominated by the residual, leading to the inefficient usage of parameters and sub-optimal performance.
    In addition, the variance of SAN results produced by ReLU is related to the sequence length which will lead to the variance exploding problem. 
    2) Regarding the normalization, since Softmax provides exponential normalization on the elements while ReLU does not, the distribution of Softmax is more centralized compared with ReLU. When dealing with a large number of value slots, Softmax restricts the utilization of more context information and leads to sub-optimal performance, while ReLU does not.
    \item When Softmax is equipped with layer normalization in key-value memory, the FFN and key-value memory are equivalent. The layer normalization can largely alleviate the scale and over-centralized problem caused by Softmax, which can boost performance.
    \item The ReLU is good at handling a large number of key-value slots (in FFN and key-value memory) and long sequences (in SAN). With quantitive analyses, we find that ReLU is less centralized, thus it can integrate the context information of more tokens when the sequence is long.
    \item As a whole, the Transformer can be viewed as the memory network, where FFN and SAN are global and local memory respectively. For FFN, the keys and values are parameters of two linear projection weights, which are globally shared by all input queries. For the SAN, the keys and values are constructed from the input sequences locally.
\end{itemize}

\section{Conclusion}
In this work, we revisit the relations between the Transformer components: the self-attention and feed-forward network, and key-value memory. Then we propose the full ReLU architecture which can achieve competitive performance with the vanilla Transformer. We have the following findings: 1) The FFN and key-value memory are equivalent when layer normalization is introduced; 2) Compared with Softmax, ReLU performs better when the number of memory value slots is large; 3) With specific designs, the proposed the full ReLU architecture can work effectively in sentence-level translation and significantly outperform vanilla Transformer in long document translation.

\section{Limitation and Future Works}
\label{app:limi}
This work has the following limitations and will be extended from several aspects. First, we only conduct experiments on the machine translation task currently.
Secondly, although we replace Softmax with a more efficient function ReLU, we obtain slight latency gains
since it is still $O(N^2)$ complexity in self-attention.
In the future, we will conduct experiments on more tasks, including language modeling, text summarization, etc. Then, since our work is parallel to the work designing the efficient Transformer, we will investigate better methods in self-attention to improve latency.

\clearpage


\bibliography{example_paper}
\bibliographystyle{icml2023}

%%%%%%%%%%%%%%%%%%%%%%%%%%%%%%%%%%%%%%%%%%%%%%%%%%%%%%%%%%%%%%%%%%%%%%%%%%%%%%%
%%%%%%%%%%%%%%%%%%%%%%%%%%%%%%%%%%%%%%%%%%%%%%%%%%%%%%%%%%%%%%%%%%%%%%%%%%%%%%%
% APPENDIX
%%%%%%%%%%%%%%%%%%%%%%%%%%%%%%%%%%%%%%%%%%%%%%%%%%%%%%%%%%%%%%%%%%%%%%%%%%%%%%%
%%%%%%%%%%%%%%%%%%%%%%%%%%%%%%%%%%%%%%%%%%%%%%%%%%%%%%%%%%%%%%%%%%%%%%%%%%%%%%%
\newpage
\appendix
\onecolumn

\section{Datasets and Implementation Details of Sentence-Level Translation}
\label{append:dataset_nmt}
\subsection{Datasets}
We evaluate our model on two widely used public sentence-level machine translation datasets: IWSLT14 De-En and WMT14 En-De, which have 153K/4.5M bilingual sentence pairs in corresponding training sets. For IWSLT14 De-En, following prior works~\cite{ranzato2015sequence,bahdanau2016actor,guo2019non}, we use 7K data split from the training set as the validation set and use the concatenation of dev2010, tst2010, tst2011 and tst2012 as the test set. For WMT14 En-De task, we use newstest2013 and newstest2014 as the validation and test set respectively. 
We use byte-pair encoding (BPE)~\cite{sennrich2015neural} to tokenize and segment all the data into subword tokens. We share the source and target vocabulary and the embedding in each language pair. And the vocabulary is extracted by Moses~\cite{koehn2007moses} for each training set with default hyperparameters.

\subsection{Implementation Details}
We follow the same encoder and decoder architecture as in Transformer~\cite{vaswani2017attention}. For the IWSLT14 De-En task, we use 6 encoder and decoder layers, 4 multi-heads, 512 and 1024 as the hidden and FFN inner dimension. 
For the WMT14 En-De task, we use 8 multi-heads and 2048 as the FFN inner dimension. 
During training, we apply dropout to the residual connections and attention weights with a rate of 0.1. We tune model parameters using Adam~\cite{kingma2014adam} ($\beta_1=0.9, \beta_2=0.98$) with label smoothing of 0.1. We schedule the learning rate following \cite{vaswani2017attention} with a warmup step of 4K. Each training batch contains around 8192 tokens, and the model is trained with 10w steps.
While inference, we set the length penalty to 1.1 for IWSLT14 De-En task and 0.6 for WMT14 De-En task. The beam size is set to 5 for all tasks. We report tokenized BLEU scores as the evaluation metric. 
Importantly, the choice of entropy upper bound $C$ in Equation (\ref{equ:normalization_loss}) should depend on the sequence length $n$ as the entropy varies on different sizes. Therefore we set $C = 0.7 log(n)$ in all experiments.

\section{Datasets of Document Translation}
\label{appendix:document_translation}
We evaluate our model on a widely used document translation dataset Europarl7 En-De~\citep{maruf2019selective,zheng2020towards}. 
To compare the models 
under different lengths of input sequences, we reconstruct the documents by concatenating sentences to different limitations of lengths. Given a length limitation $L$ and $K$ sentences with $\{l_1, l_2, ..., l_K\}$ tokens in a document, we select $m$ consecutive sentences that satisfy $\sum_1^m l_i <= L$ and $\sum_1^{m+1} l_i > L $ to form a new document. The next document will start with the sentence $l_{m+1}$. We vary the document length limitation in $L \in \{128, 256, 512, 1024, 2048\}$. We construct the validation and test sets with the same procedure as the training set.

\section{Baseline Details for Sentence-Level Translation}
\label{appendix:baseline_details_nmt}
\paragraph{Sparsemax} The Sparsemax~\cite{martins2016softmax} is a wildly used sparse activation function. Compared with Softmax, Sparsemax sets the low score elements to zero 
according to a pre-defined threshold 
to make the results sparse. 

\paragraph{1.5Entmax} The 1.5Entmax is a wild variant of the $\alpha$-Entmax sparse activation function family~\cite{peters2019sparse}, with setting $\alpha$ to 1.5. It is a more general form that includes Sparsemax ($\alpha=1$) and Softmax $(\alpha=2)$ as particular cases. 

\paragraph{ReLA} Rectified Linear Attention (ReLA)~\cite{zhang2021sparse} is an efficient sparse attention framework. Despite the Softmax activation function, they also use the ReLU function. They propose the RMS normalization with the gating mechanism to stable the training process and boost performance.

\end{document}